%% file: Final_Manuscript.tex
\documentclass[journal]{IEEEtran}
\input{Config.tex}
\begin{document}
\title{Reservoir Network with Structural Plasticity for Human Activity Recognition}

\author{
Abdullah M. Zyarah, Alaa M. Abdul-Hadi, and
 Dhireesha Kudithipudi
\thanks{
- 2024 IEEE personal use of this material is permitted. Permission from IEEE must be obtained for all other uses, in any current or future media, including reprinting/republishing this material for advertising or promotional purposes, creating new collective works, for rescale or redistribution to servers or lists or reuse of any copyrighted component of this work or other works. DOI 10.1109/TETCI.2023.3330422.

-Abdullah M. Zyarah is with the Neuromorphic AI Lab, Department of Electrical and Computer Engineering, University of Texas at San Antonio and Department of Electrical Engineering, University of Baghdad (E-mail: abdullah.zyarah@uob.edu.iq).

-Alaa M. Abdul-Hadi is with the Department of Computer Engineering, University of Baghdad (E-mail: alaa.m.abdulhadi@coeng.uobaghdad.edu.iq).

-Dhireesha Kudithipudi is with the Neuromorphic AI Lab, Department of Electrical and Computer Engineering, University of Texas at San Antonio, TX 78249 USA (E-mail: dk@utsa.edu).

}}

\IEEEoverridecommandlockouts
\IEEEpubid{\makebox[\columnwidth]{~XXXX-XXXX
\copyright2023
IEEE \hfill} \hspace{\columnsep}\makebox[\columnwidth]{ }} 

\markboth{IEEE TRANSACTIONS ON EMERGING TOPICS IN COMPUTATIONAL INTELLIGENCE}{Shell \MakeLowercase{\textit{et al.}}: A Novel Tin Can Link}

\maketitle

\begin{abstract}
The unprecedented dissemination of edge devices is accompanied by a growing demand for neuromorphic chips that can process time-series data natively without cloud support. Echo state network (ESN) is a class of recurrent neural networks that can be used to identify unique patterns in time-series data and predict future events. It is known for minimal computing resource requirements and fast training, owing to the use of linear optimization solely at the readout stage.  In this work, a custom-design neuromorphic chip based on ESN targeting edge devices is proposed. The proposed system supports various learning mechanisms, including structural plasticity and synaptic plasticity, locally on-chip. This provides the network with an additional degree of freedom to continuously learn, adapt, and alter its structure and sparsity level, ensuring high performance and continuous stability. We demonstrate the performance of the proposed system as well as its robustness to noise against real-world time-series datasets while considering various topologies of data movement. An average accuracy of 95.95\% and 85.24\% are achieved on human activity recognition and prosthetic finger control, respectively. We also illustrate that the proposed system offers a throughput of 6$\times$$\boldsymbol{10^4}$ samples/sec with a power consumption of 47.7mW on a 65nm IBM process.
\end{abstract}

\IEEEpeerreviewmaketitle

\section{Introduction}\label{sec:introduction}
\IEEEPARstart{T}{he} last decade has seen significant advancement in neuromorphic computing with a major thrust centered around processing streaming data using recurrent neural networks (RNNs). Despite the fact RNNs demonstrate promising performance in numerous domains including speech recognition~\cite{karita2019comparative}, computer vision~\cite{pang2019deep}, stock trading~\cite{bhoite2022stock}, and medical diagnosis~\cite{senturk2022layer}, such networks suffer from slow convergence and intensive computations~\cite{vlachas2020backpropagation}. Furthermore, training such networks is challenging especially when using backpropagation through time as the gradient may fade or vanish, converging the network to poor local minima~\cite{gauthier2021next}. In order to bypass these challenges, Jaeger and Maass suggest leveraging the rich dynamics offered by the networks' recurrent connections and random parameters and limit the training to the network advanced layers, particularly the readout layer~\cite{maass2002real, jaeger2001echo, liu2020quantized}. With that, the network training and its computation complexity are significantly simplified. There are three classes of RNN networks trained using this approach known as a liquid state machine (LSM)~\cite{maass2002real}, delayed-feedback reservoir~\cite{appeltant2011information, bai2018dfr}, and echo state network (ESN) which is going to be the focus of this work. 

ESN is demonstrated in a variety of tasks, including pattern recognition, anomaly detection~\cite{ullah2022intelligent}, spatial-temporal forecasting~\cite{sun2021improved}, and modeling dynamic motions in bio-mimic robots~\cite{soliman2021modelling}. Running ESN on edge devices using custom-built computing platforms yields a significant improvement in power consumption and latency~\cite{kudithipudi2016design, alomar2020efficient}. Therefore, in the literature, several research groups proposed analog, mixed-signal, and digital implementations of the ESN~\cite{yu2022performance,liang2022rotating,kume2020tuning, kleyko2020integer}. In 2020, ESN hardware implementation based on the MOSFET crossbar array is presented by Kume et al., in which the authors leverage the CMOS device variations to implement input and reservoir layers random weights~\cite{kume2020tuning}. The authors also suggest applying the circular law of the random matrix to ensure network stability and demonstrate network performance on the Mackey-Glass dataset (inference only). Then, Zhong et al. proposed a parallel dynamic mixed-signal architecture of ESN in 2021~\cite{zhong2021dynamic}. The proposed design is based on memristor devices and uses a controllable mask process to generate rich reservoir states. The design operation is validated on spoken-digit recognition and time-series prediction of the H\'enon map. 

When it comes to digital implementations, Honda et al. presented an FPGA-based implementation of ESN in 2020. To simplify computations, the authors use efficient dataflow modules and resort to ternary weight representation, resulting in a significant reduction in power consumption and area while maintaining reasonable performance. The proposed design is evaluated by predicting sine and cosine waves~\cite{honda2020hardware}. In the same year, Kleyko et al. presented an approximation of the ESN using hyperdimensional computing (HD) to implement the network efficiently in digital hardware~\cite{kleyko2020integer}. The architecture is implemented using high-level synthesis using the C programming language and used as an IP ported on Xilinx FPGA. The proposed design is benchmarked for time-series classification using univariate and multivariate datasets. Lately, Gan et al. proposed a cost-efficient (fast and energy-efficient) FPGA implementation of ESN~\cite{gan2021cost}. The proposed design takes advantage of FPGA embedded DSP units rather than configurable logic blocks to model ESN neurons. The authors trained the network offline and verified it for prediction and symbol detection tasks using the NARMA-10 dataset and software-defined radio (SDR) platform, respectively. To the best of the authors' knowledge, there is no custom-design digital implementation of ESN that considers embedded plasticity mechanisms, where reservoir space can change to improve the linear separability of the input signal. Moreover, most of the previously proposed designs lack on-chip training and network reconfigurability, critical for continuous network adaptation and stability. Offering on-chip training for ESN is vital as such a network deals with stationary and non-stationary streaming data that are impacted by noise and continuously changing. Key contributions of this work include: 
\begin{itemize}
    \item A custom-design neuromorphic chip based on ESN for human activity recognition and prosthetic finger control, targeting edge devices with stringent resources. 
    \item The proposed system supports on-chip training with dynamic sparsity and is the first of its type to use very a simplistic approach to satisfy the echo state property. 
    \item Evaluation of the system against several metrics, such as throughput, considering various of data movement topologies, robustness to noise, and power consumption including peripherals. 
\end{itemize}

The rest of the paper is organized as follows: Section II discusses the theory behind ESN and its training procedure. Section III presents the system design and implementation of the ESN. The design methodology is introduced in Section IV, while Section V presents the evaluation approach. We report on the experimental results in Section VI and conclude the paper in Section VII. 

\section{Background}
\subsection{Echo State Network}
\begin{figure} [t]
\begin{center}
\includegraphics[width=0.48 \textwidth]{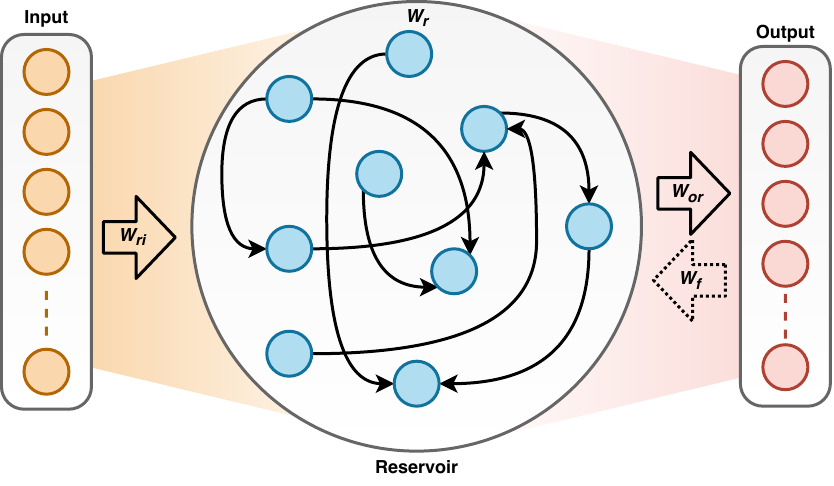}
\caption{High-level diagram of the ESN, which mainly comprises of input, reservoir, and readout layer. The input layer buffers the input, whereas the reservoir layer and readout layer stochastically extract input features and classify it, respectively.}
\label{esn}
\end{center}
\end{figure}

ESN is a class of recurrent neural networks aimed at emulating the mechanisms of sequence learning and modeling dynamical systems in the cerebral cortex. The algorithm was originally proposed by Jagear~\cite{jaeger2001echo}, and since then has received significant attention in the machine learning community, as it not only provides start-of-the-art performance in numerous applications, but also offers fast training. The general structure of the ESN, shown in~\fig{esn}, is composed of three meshed consecutive layers, namely input, reservoir (hidden), and readout (output). While the input layer is employed as a buffer, the reservoir layer, which is untrained and sparsely connected, serves two purposes. First, it acts as a memory, providing temporal context~\cite{lukovsevivcius2012practical}. Second, it works as a reservoir that projects input data into a high-dimensional feature space (non-linear expansion) ensuring linear separability of input features. The output of the reservoir layer is relayed to the readout layer, where it is linearly separated.

Mathematically, the network operations can be described as follows: given an input feature vector $\boldsymbol{u}(t)\in\mathbb{R}^{n_i \times 1}$ and a targeted output vector $\boldsymbol{y}(t) \in \mathbb{R}^{n_o \times 1}$, where $n_i$ and $n_o$, respectively, indicate the number of features per input sample and the corresponding class labels, $t = 1, ...., T$ is the discrete-time, and $T$ is the total number of training examples, the network initially computes the internal dynamic state of the reservoir $\boldsymbol{\hat{x}}(t)$. This is done via accumulating the following terms: i) input projection: dot product of input feature vector by the corresponding input-reservoir synaptic weight matrix ($\boldsymbol{W}_{ri}\in \mathbb{R}^{n_r \times n_i}$), where $n_r$ refers to the neuron count in the reservoir layer; ii) lateral projection: dot product of the reservoir output vector from the previous time step $\boldsymbol{x}(t-1)$ by the reservoir weight matrix ($\boldsymbol{W}_r \in \mathbb{R}^{n_r \times n_r}$), wherein $W_r$, most of the elements are chosen to be zeros to impose sparsity in the reservoir layer. Despite this has a marginal impact on the network performance, it enables fast reservoir updates; iii) output projection: dot product of the reservoir-output-feedback weight matrix ($\boldsymbol{W}_f\in \mathbb{R}^{n_o \times n_r}$) by the network generated output $\boldsymbol{\hat{y}}(t-1)$. The network output projection through $\boldsymbol{W}_f$ to the reservoir layer (feedback) is optional and relies on the targeted task. It is added when ESN is used in pattern generation tasks where the output can precisely be learned. Once the accumulation term is computed, it is presented to a non-linear activation function ($f$), which can either be a sigmoid or hyperbolic tangent (tanh), see~\eq{reser_eq}. Then, the reservoir neuron activation $\boldsymbol{x}(t)$ is determined as in~\eq{reser_eq2}, which is highly controlled by the leaking rate $\delta$ as ESN uses leaky-integrated discrete-time continuous-value neurons in the reservoir layer. 

\begin{equation}
\boldsymbol{\hat{x}}(t) = f[\boldsymbol{W}_{ri} \boldsymbol{u}(t) + \boldsymbol{W}_{r} \boldsymbol{x}(t-1) + \underbrace{\boldsymbol{W}_{f} \boldsymbol{\hat{y}}(t-1)}_{\text{Feedback}}]   
\label{reser_eq}
\end{equation}
\begin{equation}
\boldsymbol{x}(t) = (1 - \delta) \boldsymbol{x}(t-1) + \delta \boldsymbol{\hat{x}}(t)   
\label{reser_eq2}
\end{equation}

Once the output of the reservoir layer is determined, it is transmitted to the readout layer through output-reservoir synaptic weight matrix ($\boldsymbol{W}_{or} \in \mathbb{R}^{n_o \times n_r}$) to approximate the final output of the output as follows:

\begin{equation}
\boldsymbol{\hat{y}}(t) = f[\boldsymbol{W}_{or} \boldsymbol{x}(t)]   
\label{reser_out}
\end{equation}

\begin{figure*}[t!]
    \centering
    \includegraphics[width = 0.75 \textwidth]{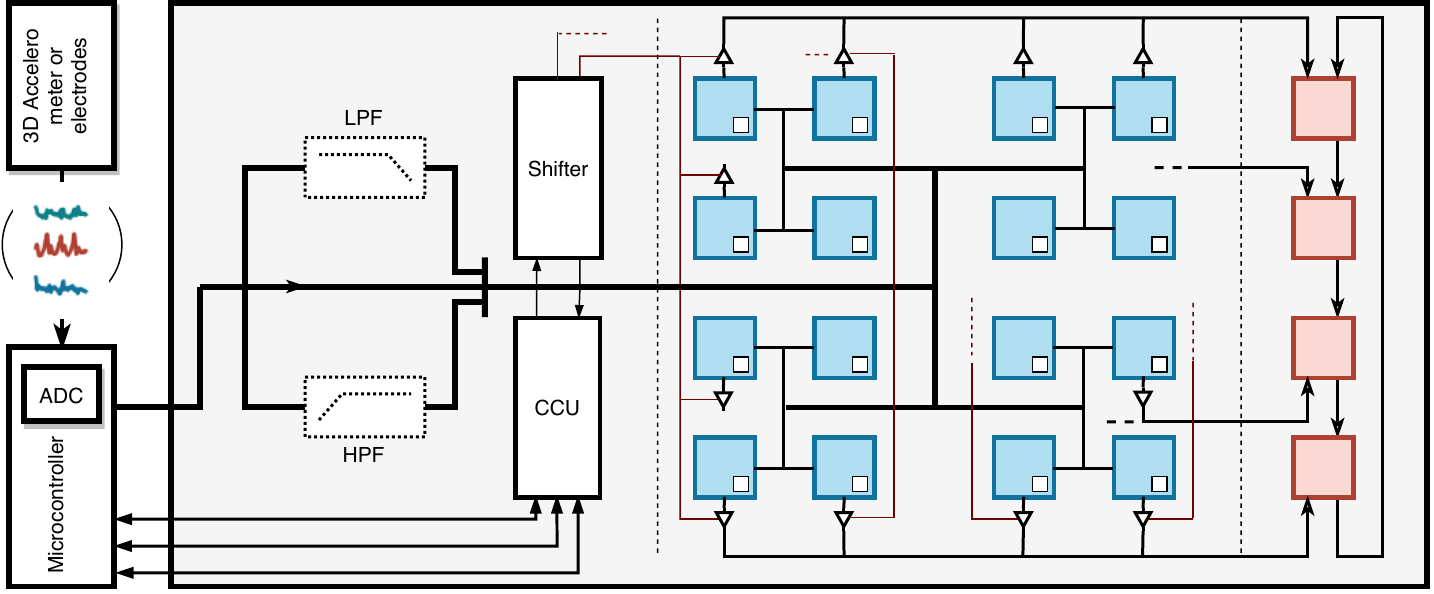}
    \caption{The system high-level architecture of the proposed ESN is mainly composed of FIR filters (optional), central control unit (CCU), shifter, and a group of neurons clustered into two layers: reservoir and readout. The CCU controls the data flow and core unit synchronization, and the shifter manages the sequential data transfer from the reservoir (blue) to output (red) neurons. The reservoir neurons provide non-linear expansion ensuring linear separability of input features by the output neurons. \vspace{-1.5mm}}
    \label{esn_system}
\end{figure*}

It is important to mention here that the ESN input vector size ($n_i$) and the readout layer size ($n_o$) are defined by the input features and the number of class labels that must be recognized. However, this is not the case when it comes to the reservoir layer whose size ($n_r$) is set randomly or experimentally based on previously provided information. Randomly, one may follow the general rule of thumb, which states that the larger reservoir produces better performance, since it is easier to find linear combinations of the signals to approximate the targeted output~\cite{lukovsevivcius2012practical}. It turns out that this approach unnecessarily increases the computing overload and the network storage requirements. Experimentally, one may start with the lowest possible reservoir size which can be set according to input data complexity and the required ESN memory capacity. Then, expand the size of the reservoir as needed. While the last approach is attractive, it is challenging to realize in hardware as it requires an ESN architecture endowed with the structure plasticity property. 

\subsection{Echo State Network Training} \label{esn_training}
ESN is well-known for its fast training, as weight adjustment is confined only to the connections that lead to the readout layer. The weights associated with the reservoir neurons are randomly initialized and left unchanged during network training. Typically, finding the optimal set of readout layer weights ensuring high network performance is done either analytically or iteratively (in both cases, it is based on the reservoir dynamic response and the ground-truth class labels). Analytically, it is done by ridge regression (normal equation) with regularization term ($\beta$) as given in~\eq{ridge}, where $\boldsymbol{X} \in \mathbb{R}^{n_i \times T}$ and $\boldsymbol{Y} \in \mathbb{R}^{n_o \times T}$. Iteratively, it is achieved using the stochastic gradient descent (SGD) given in~\eq{sgd_eq}, where $\alpha$ and $n_t$ are the learning and update rates ($n_t$ is set to 1 in this work). 

\begin{equation}
    \boldsymbol{W}_{or} = (\boldsymbol{X^TX} + \beta \boldsymbol{I})^{-1}\boldsymbol{X^T Y}
\label{ridge}
\end{equation}
\begin{equation}
    \boldsymbol{W}_{or} = \boldsymbol{W}_{or} - \frac{\alpha}{n_t} \sum\limits^{n_t}_{t=0} [\boldsymbol{\hat{y}}(t) - \boldsymbol{y}(t)] \boldsymbol{x}(t)
\label{sgd_eq}
\end{equation}

Since one of the aims of this work is to enable on-chip training, approaches based on the normal equation are avoided. The normal equation approach requires computation of the matrix inverse, which is computationally intensive and requires massive storage~\cite{zyarah2019neuromemristive}. Furthermore, it is considered as an effective solution when processing stationary data. Hence, in this work, we used SGD due to its simplicity and minimal use of compute resources when implemented in hardware. SGD also endows the system with the capability to learn from non-stationary data where the targeted output changes continuously. It is important to note that one may also resort to the recursive least squares (RLS) algorithm to train ESN. In RLS, the inverse term in \eq{ridge} is computed offline based on pre-collected data just as in ridge regression. Then, the weights are iteratively fine-tuned in the direction that minimizes the instantaneous error. Although RLS is known for its fast convergence time, it is computationally more intensive than SGD.

\begin{figure*}[h!t]
\begin{center}
\includegraphics[width=0.8 \textwidth]{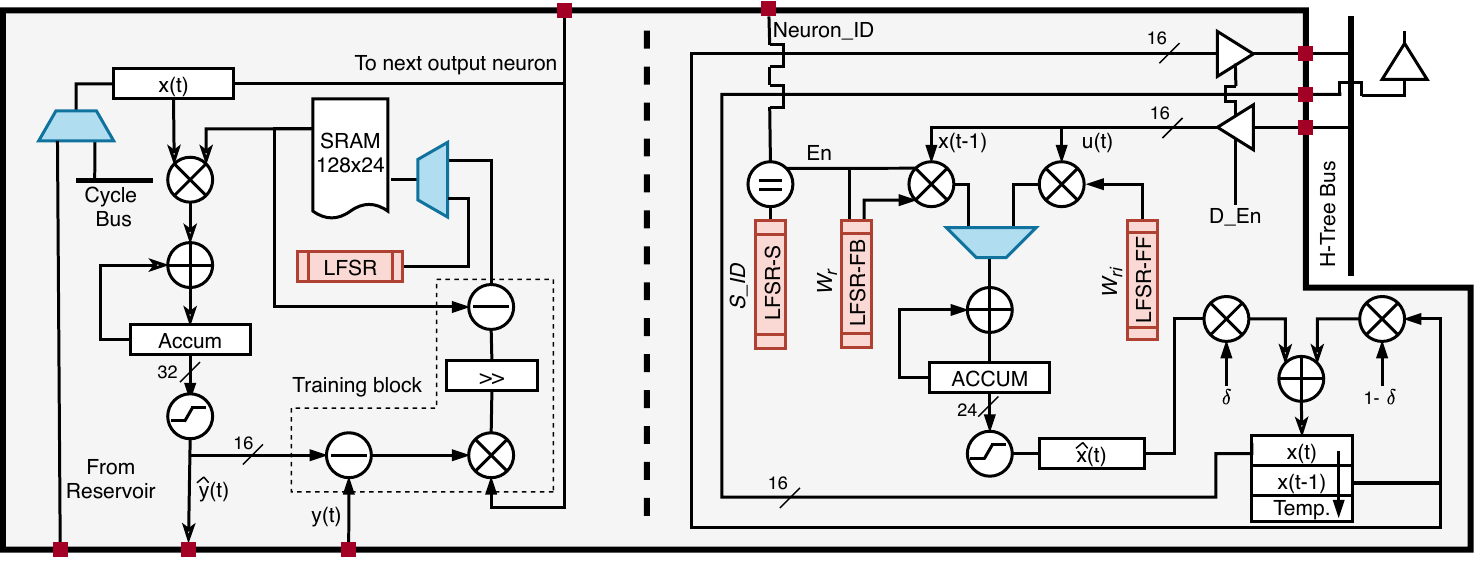}
\caption{The RTL schematic of the: (Left) Output neuron, in which the weight-sum of the reservoir output is computed and the weight adjustment is taking place; (Right) Leaky-integrated discrete-time continuous reservoir neuron used in the proposed ESN system.\vspace{-2.5mm}}
\label{hidden}
\end{center}
\end{figure*}
\section{System Design and Implementation}
The high-level architecture of the ESN-based system is shown in~\fig{esn_system}. The system is running in a synchronous fashion, where input signals are initially sampled, digitized, filtered, and then presented to the network for feature extraction. The filtering here is optional and can be used to boost the performance of the network when a small reservoir is used. It is realized using a set\footnote{The set is equal to the number of input features.} of third-order low-pass filters (LPFs) and high-pass filters (HPFs) with a cutting frequency of 1Hz. All filters are approximated by finite impulse response filters (FIR), which can be custom designed or instantiated as an IP. The outputs of the input layer are multiplied by their corresponding random weights in the reservoir layer. Once the feedforward weighted inputs of the reservoir neurons are received, collectively with weighted reservoir neuron responses from the previous time step and the projected weighted output feedback (optional), the non-linear responses of the reservoir neurons are determined. The output of the reservoir is classified using the readout layer, which is trained online as alluded to earlier in section~\ref{esn_training}.

\begin{figure}[h!b]
\begin{center}
\includegraphics[width=0.35 \textwidth]{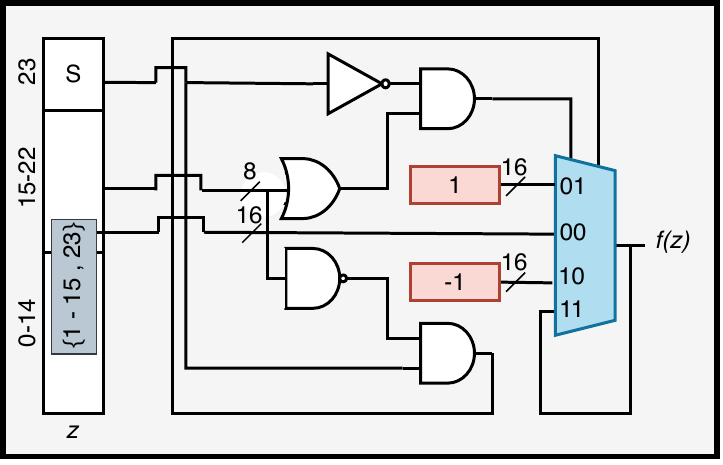}
\caption{RTL design of hyperbolic tangent ($tanh$) activation function modeled using piece-wise functions.}
\label{tanh_fig}
\end{center}
\end{figure}

Given a set of input feature vectors (unfiltered) and their corresponding class labels $\boldsymbol{S}(t) \!=\{(\boldsymbol{u}(t), \boldsymbol{y}(t)), (\boldsymbol{u}(t+1), \boldsymbol{y}(t+1)), ...\}$, the features are initially quantized using signed fixed-point notation of 1 sign bit, 3 integer bits, and 12 fractional bits (SQ3.12). Then, the quantized inputs are presented to the network via the microcontroller when it is triggered by the central control unit (CCU), specifically when the network is not busy and ready to process new inputs. The presented features are broadcasted sequentially through the H-Tree network to all the reservoir neurons, which are arranged in a 2D array. In the reservoir neurons, the features are multiplied by the random weights ($\boldsymbol{W}_{ri}$) generated using linear-feedback shift register (LFSR-FF)\footnote{The MSB of the generated random numbers is reserved for sign, such that bipolar weights are achieved.}~\cite{zyarah2016optimized}. The same is applied to the responses of the reservoir neurons\footnote{Neuron activations transferred through feedback connections are randomly selected via LFSR-S to impose sparsity.} from the previous time step $\boldsymbol{x}(t-1)$, which are multiplied by random weights ($\boldsymbol{W}_r$) generated by LFSR-FB with different seeds, illustrated in~\fig{hidden}-right. The multiplication is carried out using a 16-bit fixed-point multiplier, as it consumes 5$\times$ and 6.2$\times$ less energy compared to 32-bit fixed-point and floating-point multipliers, respectively~\cite{han2016eie}. Additionally, using LFSRs rather than memory to generate random weights results in $20.1\%$ reduction in overall power consumption and $2.38\%$ reduction in system's area footprint. However, the weighted-sums of both, input features and neurons' responses, are accumulated and then subjected to a hyperbolic tangent ($tanh$) non-linear activation function to generate the output $\boldsymbol{\hat{x}}(t)$. Realizing a high-precision non-linear function approximator for $tanh$ (or even $sigmoid$) activation function requires complex mathematical operations; thus, a piece-wise model (see~\eq{tanh} and \eq{sigmoid}) is adopted to achieve efficient design in terms of resources (see~\fig{tanh_fig}). Since all reservoir neurons are leaky-integrated discrete-time continuous, the output of the reservoir neurons $\boldsymbol{x}(t-1)$ and the update rate $\boldsymbol{\hat{x}}(t)$ are controlled by leakage terms, i.e. $\delta$ and $1 - \delta$ which are stored in the CCU and shared with all reservoir neurons. Having leaky-integrated reservoir neurons is essential when dealing with streaming data, as it controls the speed of network adaptation (short-term memory).

The output of the reservoir layer is propagated to the readout layer via unidirectional links in a sequential fashion, one column at a time as controlled by the shifter. Once it is received, it is stored into internal registers and later forwarded to neighboring neurons to fulfill the dense-connection requirement. Typically, when the shifter activates one column, it activates its transmission gates and deactivates those connected to other neurons within the layer. Once the transmission gates are activated, the selected column is kept enabled until its output is processed by the readout layer. 
 
\begin{equation}
tanh(z) = 
\begin{cases}
1,~~~~~~~~~~~~~\text{if}~ z > 1\\
-1,~~~~~~~~~~\text{if}~ z < -1\\
z , ~~~~~~~~~~~~Otherwise
\end{cases}
\label{tanh}
\end{equation}

\begin{equation}
sigmoid(z) = 
\begin{cases}
1,~~~~~~~~~~~~~~\text{if}~ z > 2\\
0,~~~~~~~~~~~~~~\text{if}~ z < -2\\
\dfrac{z}{4} + 0.5 , ~~~Otherwise 
\end{cases}
\label{sigmoid}
\end{equation}

\begin{figure*}[h!t]
\begin{center}
\includegraphics[width=0.95 \textwidth]{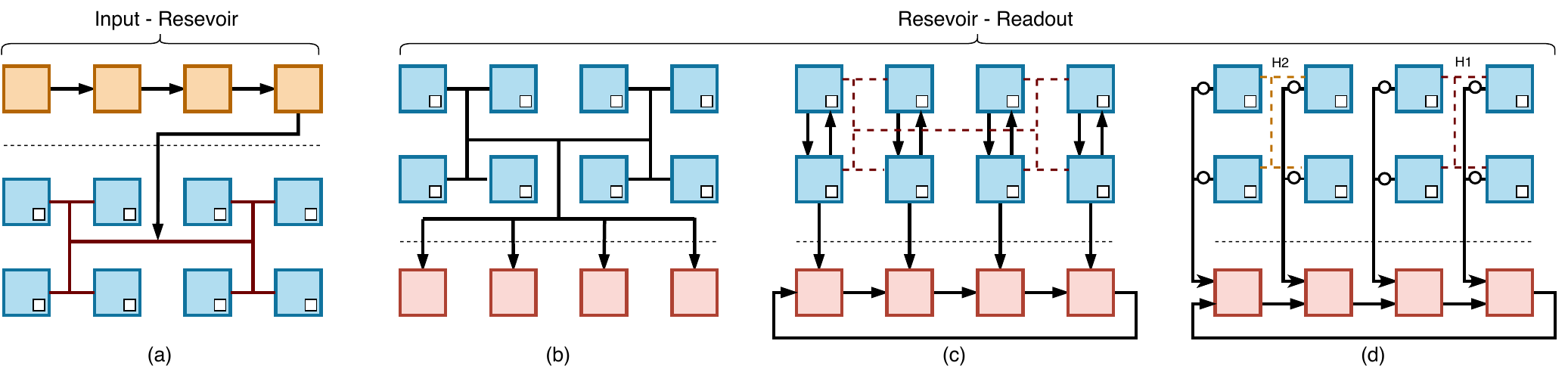}
\caption{(a) Data movement from the input layer to the readout layer via an H-Tree network, (b) SH-Tree topology which uses a single H-Tree network shared among all the PEs in the network, (c) local rings topology, in which data within the layers are circulated through separated local rings, (d) MH-Tree topology, where multiple H-Tree networks are used to move data within the reservoir and dedicated unidirectional links to transfer reservoir activation directly to the readout layer.}
\label{data_movement}
\end{center}
\end{figure*}

In the readout layer, neurons run in two modes: training and testing. In training mode, the outputs of the reservoir are multiplied by their corresponding 24-bit weights stored in distributed local SRAMs\footnote{All the initial random weights stored in SRAMs are generated using dedicated LFSRs located in output neurons. This takes place when we run the ESN system for the first time.} rather than external DRAMs to achieve more energy saving (SRAM --- 32-bit 32KB in 45nm process --- access consumes 128$\times$ less energy than DRAM~\cite{horowitz20141}). Then, the weighted-sum is presented to the sigmoid activation function to generate the final network response for a given input, see~\fig{hidden}-left. The network response is also passed to the training block to compute the gradient since the network is trained with SGD. The gradient is computed by estimating the network error, comparing network response and the ground-truth label, see~\eq{sgd_eq}. The resulting network error is multiplied by the corresponding reservoir output and the learning rate. Here, the multiplication by the learning rate is carried out by means of a right-shift operation to reduce the number of used fixed-point multipliers. Once the gradient is computed, it is subtracted from the weights. The new weights overwrite the old weight values stored in SRAM and the process is repeated until all weights are updated. According to the experimental results, the value of the gradient range is (1,5e-7] and the weight range after training is [-5,5]. This implies that the readout layer weight precision should be no less than 24 bits arranged as SQ3.21, which can handle gradient values as low as 4.76e-7 ($\equiv$1/$2^{21}$). During testing, the same procedure is followed except that no gradient is computed nor weight adjustment occurs.


It is important to mention here that there are points in time when the ESN network needs to interact with the external microcontroller for network hyper-parameters adjustment. The adjustment is limited to tuning the leakage rate, learning rate, and reservoir size within pre-defined ranges. However, this occurs when the ESN network shows unsatisfactory performance. 



\section{Design Methodology}
\subsection{Network parameters}
In ESN, performance is highly impacted by network hyperparameters and the statistical nature of the time-series data. Thus, in this work, we endowed our system with the ability to tune four key hyperparameters that unequally contribute to ESN performance, throughput, and energy efficiency. First, the size of the reservoir, which as a rule of thumb, the bigger ensures better performance because it increases the separability of the extracted features. However, due to the limited resources of edge devices, one can rely on the approach suggested by~\cite{jaeger2002short}, which states that the size of the reservoir can be set according to the number of real values remembered by the network to successfully accomplish a given task. This demands an earlier familiarity with the data being processed, which makes it an unpractical approach. Thus, structural plasticity, particularly neurogenesis, plays an important role here, as it endows the network with the ability to linearly add more reservoir neurons. According to our reported results and those published in previous works~\cite{kudithipudi2016design, kume2020tuning}, one can initially set the reservoir size at 100 to achieve satisfactory performance and expand the size linearly as needed. Second, the reservoir sparsity, which may slightly enhance the performance, also leads to substantial improvement in computational speed and energy saving. Experimentally, a sparsity of less than 10\% of the reservoir size always results in improved performance and rapid reservoir update. Third and last are the leakage and learning rates, which control the reservoir dynamic update and the ESN convergence speed, respectively. Therefore, these parameters can be continuously updated, especially, when ESN is involved in multiple tasks. It should be highlighted here that setting the aforementioned parameters is done via the particle swarm algorithm (PSO)~\cite{eberhart1995new} with 50 particles running on the external microcontroller. 

\subsection{Data Movement} \label{data_mov}
The data movement here refers to the approach followed to transfer the activations of the reservoir neurons within the layer, reservoir-to-reservoir (RTR), and outside it, reservoir-to-readout (RTO). Typically, this is controlled by topology of the interconnection network, which dictates the connection pattern of the neurons within the network. Although several interconnection network topologies for the ESN algorithm have been presented in literature~\cite{kudithipudi2016design, gallicchio2019reservoir} they are solely limited to the reservoir layer. In this work, we will discuss RTR and RTO data movement and investigate its impact on ESN performance. Then, we will characterize each topology by coverage and serialization latency ($\ell$)\footnote{In this work, we defined the serialization latency as the time required to broadcast the reservoir activations carried by the feedback connections added to the time required to transfer them to the readout layer.}:


\begin{itemize}
    \item SH-Tree: In this topology, shown in~\fig{data_movement}-(b), a single H-Tree network is used to connect all the neurons in the network; therefore, the sequential nature of data movement dominates here. Specifically, when the neuronal activations are moved from the previous time step within the reservoir and the final output of the reservoir neurons is broadcast to the readout layer. Although this topology offers full coverage and uses fewer interconnects, it is known to be slow, as using an H-Tree network that is shared across layers hinders the possibility of pipelining phases to speed the computations and maximize the resource utilization. The serialization latency of this topology, given a network size of 4x128x4 with 16-bit channel and 10\% reservoir sparsity ($\frac{1}{\kappa}$), is 1408 cycles, as estimated by \eq{t2}.
    
    \begin{equation}
        \ell = n_{r} (\lfloor{\frac{n_r}{\kappa}}\rfloor +1)
        \label{t2}
    \end{equation}

    \item Local rings: In this topology, shown in~\fig{data_movement}-(c), the reservoir neurons are arranged into a 2D array connected to the input layer through an H-Tree network and have a local ring-shaped interconnections network within the layer. When the reservoir neurons compute their internal dynamic states, the activations from the previous time step are cycled row-wise via the local rings. Then, the final output of the reservoir neurons is propagated row-wise to the readout layer. In the readout layer, the activations are cycled within the layer to compute the weighted sum of each output neuron. Although this topology is known to reduce data movement because the transferred activations can be temporarily stored and used later when computing the internal dynamic state of the reservoir neurons, it comes at the expense of increasing the temporary storage of the individual neurons. Besides, this topology allows the feedback connections to grow only in local regions within the reservoir, which negatively affects the performance of the network by (3-4)\%. The serialization latency (worst case scenario), can be computed using~\eq{t1}, where $n_{row}$ represents the number of rows in the 2D reservoir array (32$\times$4). For the 4x128x4 network size, the serialization latency, given a 64-bit (4$\times$16) channel, is estimated to be 480 cycles.
    \begin{equation}
        \ell = n_{row} (\lfloor{\frac{n_r}{\kappa}}\rfloor + n_o -1)
        \label{t1}
    \end{equation}
    
    \item MH-Tree: Multiple H-Tree networks are utilized in this topology to receive input features and to broadcast reservoir activations via feedback connections. Using multiple H-Trees reduces the latency required to broadcast reservoir activations via feedback connections, and also the capacitance load with a slight improvement in performance. Unlike the previous approach, the reservoir neurons are not pipelined, rather they can transfer data directly to the readout layer. This replaces temporary storage with tri-state buffers, which eventually minimizes the energy consumption and footprint area. Furthermore, the activations of the reservoir neurons are transferred to the readout layer via separated buses rather than the H-Tree. This endows the ESN with the capability of pipelining phases and avoids unnecessary delay. Serialization latency (without pipelining) is estimated using~\eq{t3}. For the network size as in the above examples, it is estimated to be 864 cycles, such that ($\frac{{n_r}}{\sigma} \geq 64$), where $\sigma$ is the number of H-Trees. One may notice that the MH-Tree is 1.8$\times$ slower than the local rings. This is because the local rings topology has the advantage of simultaneously broadcasting reservoir states via feedback connections. 
    
     \begin{equation}
        \ell = \lfloor{\frac{{n_r}^2}{\sigma \kappa}}\rfloor + n_{rows} (n_o -1)
        \label{t3}
    \end{equation}  
\end{itemize}
\begin{figure}[h!b]
\begin{center}
\includegraphics[width=0.4 \textwidth]{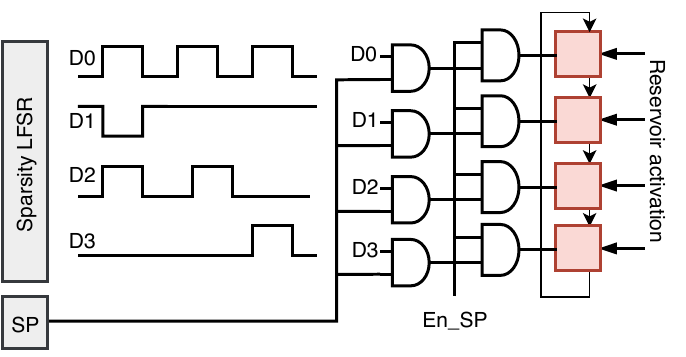}
\caption{RTL design to enable sparsity in reservoir-readout layers. The sparsity level and pattern are controlled by the LFSR and the $SP$ signal generated by the CCU.}
\label{en_sparsity}
\end{center}
\end{figure}

\subsection{Time-Multiplexed Resources}
Enabling resource sharing, where an individual neuron circuit functionally serves multiple neurons, is widely adopted in the designs of neuromorphic system~\cite{han2016eie, chen2019eyeriss}. While this approach can reduce the used resources multiple times, it comes at the expense of degrading system throughput and increasing temporary storage~\cite{zyarah2017resource}. However, such a feature can be leveraged to enable the neurogenesis aspect of structural plasticity, where neurons in the reservoir layer can be linearly adjusted. In this work, the possibility of increasing the neuron count in the reservoir layer is endowed to the designed system via the incorporation of temporary storage to preserve the internal states of the neurons. The weight values of synaptic connections and the locations of newly generated neurons with the input and within the reservoir are obtained by reusing the LFSRs equipped with each neuron, but with different seed values or slightly modified tapping. In the case of the synaptic connections with the readout layer, one may either use extra storage to account for the newly added neurons or switch from dense to sparse connections mode. The latter is adopted in this work, as it can be easily implemented and does not require additional storage.~\fig{en_sparsity} illustrates the RTL design of the circuit used to impose sparsity in the readout layer. It leverages the random sequence of binary bits generated by LFSR to randomly accept or reject data projected from the reservoir layer. The sparsity level is controlled by a sequence generated by the $SP$ signal generated by the CCU.

\subsection{Echo State Property and Chaotic Measure}
One of the crucial conditions for the ESN network to work at the edge of the chaos regime implies fulfilling the echo state property condition ($\psi(\boldsymbol{W}_r) < 1$)~\cite{legenstein2007edge, lukovsevivcius2012practical}. This condition can be satisfied by scaling the width of the distribution of the non-zero elements of the reservoir weight matrix via a global parameter known as the spectral radius $\psi(\boldsymbol{W}_r)$. Mathematically, this is carried out by computing the maximal absolute eigenvalue ($\psi$) of the reservoir weight matrix ($\boldsymbol{W}_r$) and then dividing $\boldsymbol{W}_r$ by $\psi$~\cite{lukovsevivcius2012practical}. Performing such operations in hardware involves numerous complex computations and memory access if the reservoir weights are stored in SRAMs or DRAMs. However, such a challenge does not manifest itself here. Using the LFSRs to generate random weights and enforce sparsity results in an eigenvalue that is almost the same for a fixed sparsity level regardless of the randomness. Furthermore, a  linear scaling in the sparsity level leads to an almost equal increase in the maximal absolute eigenvalue, as shown in~\fig{accur_th}-(a). Thus, satisfying the echo state property here can be done by only shifting the weights generated by LFSRs to the right, and this shift level is pre-defined by the sparsity level.~\fig{accur_th}-(a) illustrates the eigenvalues of the reservoir weight matrices as a function of sparsity level averaged over 20 runs.

Regarding the chaotic measure of the reservoir, it is estimated using the Lyapunov exponent ($\lambda$)~\cite{schrauwen2008computational}. Lyapunov exponent represents the difference in reservoir states in response to small changes in input patterns. Typically, it can be a positive value indicating a chaotic system or a negative value representing a stable system. As aforementioned, the optimal performance of the reservoir network occurs at the edge of chaos; thereby, a near-zero Lyapunov exponent is always preferred. This can be achieved when the ratio of the distance between the input vectors and the corresponding reservoir states is one~\cite{gibbons2010unifying}, see~\eq{lb_eq}. $N$ and $k$ are the total number of testing samples and a constant scaling factor, respectively. $\boldsymbol{u}_j(t)$ is the input feature at time step $t$ and $\boldsymbol{u}_{\hat{j}}(t)$ is the nearest neighbor to $\boldsymbol{u}_j(t)$. $\boldsymbol{x}_j(t)$ and $\boldsymbol{x}_{\hat{j}}(t)$ are the dynamic states of the reservoir neuron in response to $\boldsymbol{u}_j(t)$ and $\boldsymbol{u}_{\hat{j}}(t)$.~\fig{accur_th}-(b) illustrates the chaotic level of both the ESN software model and the corresponding hardware implementation, in which all weights are generated using LFSRs. While the software model seems to have a better chaotic response as compared to the hardware implementation, the difference is still marginal. The figure also shows that the chaotic response is slightly affected by the size of the reservoir, as seen in~\cite{kudithipudi2016design}. 
\begin{equation}
\label{lb_eq}
\lambda(t) \sim k \sum^{N}_{n=1} \ln \biggl(\frac{|| x_j(t) - x_{\hat{j}}(t)||}{||u_j(t) - u_{\hat{j}}(t)||}\biggl)
\end{equation}

\section{Evaluation Methodology}
\subsection{RTL Design and Physical Layout}
\begin{figure}[h!b]
\begin{center}
\includegraphics[width=0.35 \textwidth]{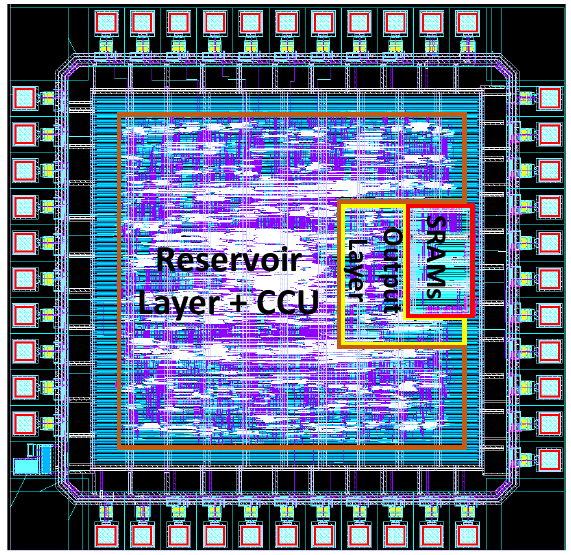}
\caption{The physical layout of the proposed ESN-based chip under IBM 65nm process. The total die area is 5.175$mm^2$}\vspace{-3mm}
\label{layout}
\end{center}
\end{figure}

In order to validate the operation of the proposed RTL design, the architecture is modeled in Python while considering the hardware constraints. This includes: i) the limited precision of weights and mathematical and logical operations; ii) piece-wise modeling of activation functions; iii) the sequence of data flow and operations. The results of the Python model are collected and compared with those of the hardware counterpart, implemented in Verilog HDL. The proposed hardware design is synthesized using the Synopsys Design Compiler (DC) under the IBM 65nm process node. Then, the design is placed and routed in Cadence Innovus (see~\fig{layout}) and optimized for power, performance, and area (PPA). When it comes to estimating the power consumption, it is done after simulating the RTL design in Cadence Simvision to record the switching activities. The power consumption of the full-chip is estimated by using Voltus, the IC power integrity solution developed by Cadence.

\begin{figure*}[h!t]
\centering
\subfigure{\includegraphics[width=59mm, height=50mm]{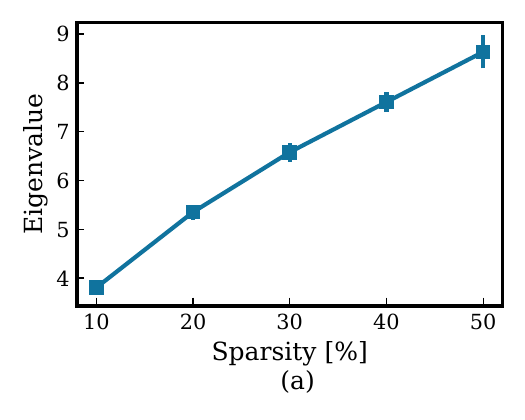}}
\subfigure{\includegraphics[width=59mm, height=50mm]{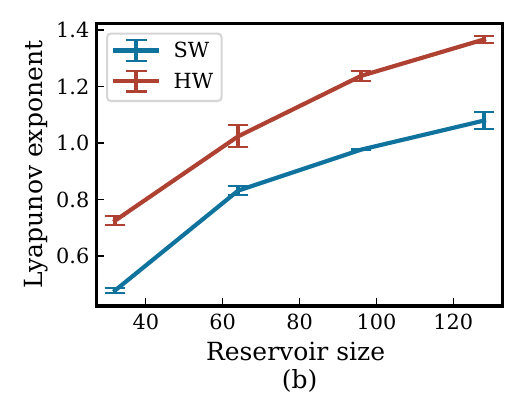}}
\subfigure{\includegraphics[width=59mm, height=50mm]{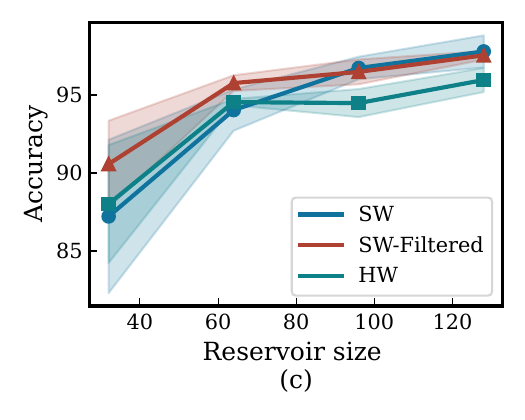}}

\subfigure{\includegraphics[width=59mm, height=50mm]{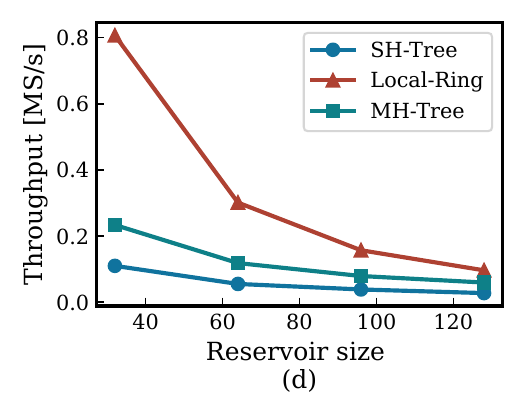}}
\subfigure{\includegraphics[width=59mm, height=50mm]{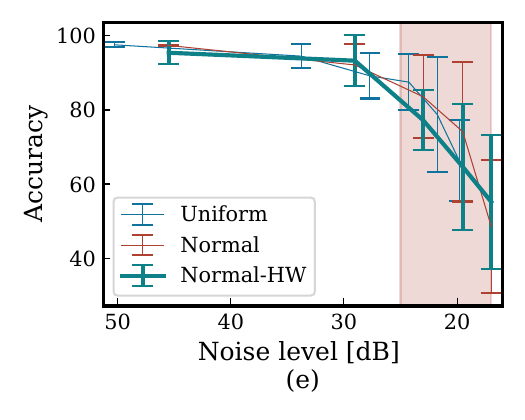}}
\subfigure{\includegraphics[width=59mm, height=50mm]{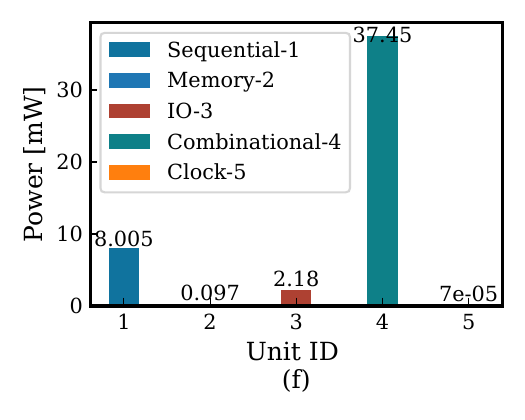}}
\caption{(a) Eigenvalues of the reservoir weight matrix as a function of sparsity level averaged over 20 runs, (b) Chaotic measure of the software and hardware ESN models using Lyapunov exponent, (c) The classification accuracy of the human activities collected using a wearable uncalibrated device mounted on the chest using ESN software model (filtered and unfiltered) and hardware model, (d) Throughput of the ESN as a function of the reservoir size while considering different topologies of data movement, (e) The ESN network classification accuracy in the presence of various degrees of noise, (f) The breakdown of the average power consumption of the designed ESN chip by component type.\vspace{-2.5mm}}
\label{accur_th}
\end{figure*}

\subsection{Benchmark}
The proposed system is benchmarked using the human activity recognition (HAR) dataset~\cite{casale2012personalization} and the prosthetic finger control dataset~\cite{khushaba2012toward}. The HAR dataset comes with more than 1,927,897 samples, split into 70\% for training and 30\% for testing. Each sample has three attributes collected from a wearable uncalibrated accelerometer mounted on the chest, particularly x-acceleration, y-acceleration, and z-acceleration. Data are collected from 15 subjects while performing 7 activities (the sampling frequency of data collection is 52Hz). Although some of these activities are distinguished, such as working at a computer, standing, walking, and going up/down stairs, others are mixed up and labeled under the same classes. For example, the following activities are recorded consecutively and labeled the same: standing up, walking, and going up/down stairs. Such classes are excluded from the training/testing phases as they may lead the network into total chaos and unfair evaluation. When it comes to the prosthetic finger control (PFC) dataset, it has EMG signals of finger movements recorded from healthy subjects. The total number of subjects is 8 (6 males and 2 females) aged between 20 and 35 years old. Signals are recorded using two EMG channels (Delsys DE 2.x series EMG sensors) placed over the circumference of the forearm and mounted via an adhesive skin interface. The recorded EMG signals are amplified using a Desys Bagnoli-8 amplifier with a gain of 1000, sampled at 4000Hz, and digitized via a 12-bit analog-to-digital converter. All signals are band-filtered between 20Hz and 450Hz to remove the 50Hz line interface. 10 classes of individual and combined finger movements were implemented by the participants and only 4 are used in this work. Each movement is repeated and recorded six times. Data collected from the first 5 trials are used for training (1,920,000 samples), whereas the last one is used for testing (384,000 samples). 

It is worth mentioning that the HAR is an unbalanced dataset, unlike the PFC dataset. Therefore, we report the classification accuracy for both datasets, but the F1 score is confined to HAR. In addition to system evaluation for classification, the HAR dataset is used to evaluate the robustness of the system to noise, throughput, and power consumption.

\begin{figure*}[h!t]
\begin{center}
\includegraphics[width=0.9 \textwidth]{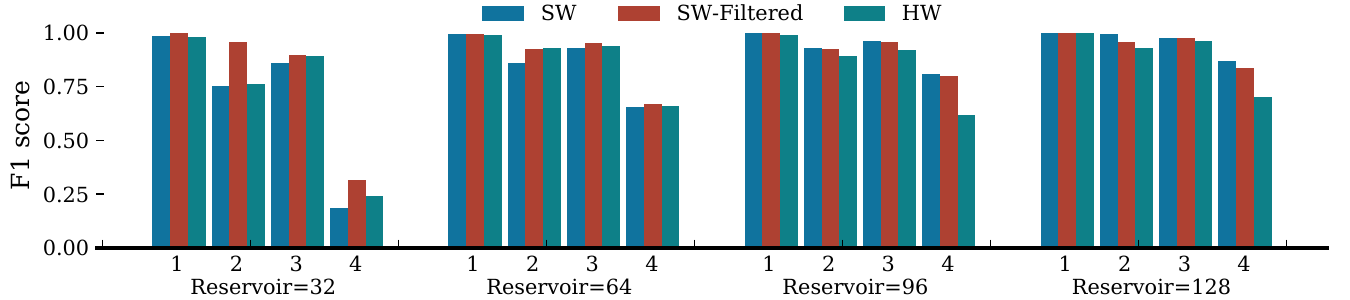}
\caption{F1 score of the ESN models, software model (filtered and unfiltered), and the hardware, when benchmarked using human activity recognition with four classes (1:working
at a computer, 2:standing, 3:walking, and 4:going up/down
stairs).}
\label{f1_scores}
\end{center}
\end{figure*}
\section{Results and Discussion}
\subsection{Classification Accuracy}
Quantifying the performance of the proposed system when processing streaming data via observing the classification accuracy can be misleading. This is mainly attributed to two reasons. First, the features associated with the classes that need to be recognized are captured over various time periods. Second, the classification accuracy is measured as the number of times the output is correctly predicted to the overall predictions made up by the network. As a result, one may get state-of-the-art performance even if the network misclassifies several classes recorded over a short period of time. Therefore, in this work, we are reporting the classification accuracy along with the F1 score, harmonic mean of precision and recall. 

\fig{accur_th}-(c) demonstrates the HAR classification accuracy of the hardware implementation as a function of the reservoir size using the proposed ESN system. It can be seen that increasing the reservoir size has a major impact on the ESN performance when classifying time-series data. This is because the kernel quality~\cite{moon2019temporal,legenstein2007edge} (a measure of linear separation property) improves as the size of the reservoir increases, reflecting an enhancement in the complexity and heterogeneity of the nonlinear operations carried out by the reservoir neurons. The maximum averaged accuracy achieved is 95.95$\pm$0.78\%. When compared with the software (filtered and unfiltered) implementation, one may observe that there is a degradation in performance by $\sim$1.8\%. The reason behind the minor drop in performance is the use of low-precision synaptic weights and the simple modeling of activation functions. A similar pattern is noticed when recording the F1 score, illustrated in~\fig{f1_scores}. The small-sized reservoir results in low performance, where two classes are partially misclassified. While this poor performance can be rectified if filters are used. However, this seems to have an effect only when using a small reservoir. When classifying finger movements, an average accuracy of 85.24$\pm$2.31\% is achieved and $\sim$1.2\%  performance degradation is observed compared to the software counterpart. Also, the same impact of reservoir size on network performance is witnessed. In addition to the size of the reservoir, changing or swapping between the recognized classes has a minor impact on the performance of the network, leading to $\sim\pm$2.1\% fluctuation in accuracy.


\subsection{Throughput}
Throughput, defined as the number of processed samples per second, is estimated for the developed ESN system while considering various paradigms of data movement.~\fig{accur_th}-d shows the change in ESN throughput as a function of reservoir size and for different topologies of data movement. It can be observed that increasing the number of neurons in the reservoir negatively impacts the throughput because of the high latency results from moving contextual data (activations via feedback connections). Besides the reservoir size, the throughput is also highly affected by the utilized topology of data movement due to serialization latency, estimated in Section~\ref{data_mov}. For instance, in MH-Tree, for a reservoir layer with 128 neurons, the throughput is enhanced by 2.14$\times$ as compared to SH-Tree. This is mainly attributed to the pipelining in the training phase and data movement phase, which itself is sped up twofold. Furthermore, the SH-Tree uses one shared H-Tree network to: i) transfer activations belonging to the reservoir neurons to the readout layer; ii) transfer activations within the reservoir layer itself, aiming to replicate the recurrent connections. This process occurs sequentially and imposes a significant delay. One may resort to reducing the delay by only broadcasting the activation of neurons with significant impact\footnote{Such approach is adopted in our work, but the evaluation is taken place outside the neurons. This is just to save resources as only one circuit is needed to evaluate the activation of all neurons broadcasted via feedback connections.}, i.e. has non-negligible output, or alternating the broadcasting of reservoir neuron activations through time to reduce the data movement by half. While the first suggestion demands more resources to evaluate each neuron output, the second degrades the network performance (classification) by 3\%. Therefore, in this work, we explored two approaches, both giving almost identical results when it comes to classification accuracy. The first approach involves using multiple H-Trees, which we have already discussed in the Data Movement section. This approach includes splitting the reservoir neurons into two clusters, the first of which is forbidden from developing outgoing recurrent connections. The second cluster is allowed to have both. These approaches enhance throughput by 1.32$\times$ and have a negligible impact on network performance ($\pm$0.2\% change in classification accuracy).

It is important to mention here that the real-time processing of the proposed ESN system is contingent by the sampling frequency of the processed time-series data and the system throughput. In this work, the time-series data of HAR and PFC are sampled at 52Hz and 4000Hz, respectively. Clocking the system at 50MHz and reducing data movement between and within the layers satisfy the throughput requirement and achieve 0.06M samples per second.


\subsection{Noise Robustness}
In order to investigate the ESN system's robustness to noise while processing streaming data, we superimposed the HAR dataset with uniform and Gaussian noise of varying degrees. Here, the amount of noise is defined by the signal-to-noise ratio (SNR) and is injected based on the Bernoulli distribution with a probability of 0.5.~\fig{accur_th}-e demonstrates the recorded classification accuracy of the noisy input, which at a glance seems to degrade gracefully for both uniform and Gaussian noise until the SNR hits 26dB. Then, the network's capability in handling the noise is degraded and starts to experience a severe drop in performance. Typically, it is preferred to maintain an acceptable performance level when dealing with noisy input as long as the SNR is more than 25dB~\cite{zyarah2020energy}, not shaded region in~\fig{accur_th}-e. Therefore, the developed ESN system manifests reasonable robustness to noise.


\begin{table*}[h!tb]
\caption{A comparison of the proposed ESN system with previous work. One may note that these implementations are on different substrates, thereby this table offers a high-level reference template for ESN hardware rather than an absolute comparison.}
\label{HardwareAnalysis}
\setlength\tabcolsep{3 pt}
\begin{center}
\begin{threeparttable}
\begin{tabular}{|c|ccccc|}
\hline                      
\rowcolor{Gray} 
\textbf{Algorithm} & \textbf{M-ESN \cite{kudithipudi2016design}\tnotex{tnote:robots-a2}} & \textbf{ESN-FPGA \cite{honda2020hardware}} & \textbf{ESN-DSP~\cite{gan2021cost}} & \textbf{Parallel-ESN}~\cite{alomar2020efficient} & \textbf{This work} \\ \hline 
  Task   & Classification    & Prediction & Prediction   & Prediction & Classification  \\ 
  Operating Frequency  & 26MHz & 200MHz &  100MHz  & 50MHz & 50MHz\\ 
  Reservoir size  & 30 & 100 & 20-100 & 200 & 128\\ 
  Input x Output size  & - & 2x2 & 1$\times$1 & 1$\times$1 & 4$\times$4\\ 
  Quantization & 30 FXP & 3 Ternary & 16-20 FXP & 16 FXP & Mixed-FXP\tnotex{tnote:robots-a3}\\
  Averaged consumed power & 47.76mW  & 670mW & 0.283W & $<$1.5W & 47.74mW\\ 
  Dataset &	ESD \&  & Sine \& & SDR \& & Santa Fe~\cite{weigend1993results} & HAR \&\\
  & PF & Cosine & NARMA-10 && PFC \\
  Latency ($\mu$s) &  -      &       200       &  0.819  & 3 & 17.36   \\ 
  Technology node &  PTM 45nm  & - &  -  & -  & IBM 65nm   \\ 
  Training & Off-Chip & Off-Chip & Off-Chip & Off-Chip & On-Chip\\
  Platform & FPGA  & FPGA & FPGA & FPGA & ASIC \\ \hline
  \end{tabular}
      \begin{tablenotes}
      \item\label{tnote:robots-a2} In~\cite{kudithipudi2016design}, the authors present memristor-based mixed-signal implementation and pure CMOS digital implementation. To make a fair comparison, only the digital implementation results are reported. The power consumption is estimated only for the hidden layer using Synopsys Primetime PX.
      \item\label{tnote:robots-a3} FXP: fixed-point representation.
    \end{tablenotes}
\end{threeparttable}
\end{center}
\vspace{-5mm}
\end{table*}

\subsection{Power Consumption}


The average power consumption of the ESN chip when performing local training is estimated to be 47.74mW.~\fig{accur_th}-f illustrates the breakdown of the average power consumption. One may notice that a small amount of power is directed towards the memory due to its small size and limited usage, unlike the sequential and combinational units that are frequently used during training and testing phases. While most of the operations in the training and testing phases are overlapped, the training phase consumes more power because it is not limited to processing input data, but also involves continual adjustment of pathway of the synaptic connections and their weight values. The proposed chip rarely runs in a third phase, namely the setup phase. The setup phase is utilized when the chip is run for the first time and a severe drop in network performance is experienced. In the setup phase, the chip core units will run in a sleep-mode until the external microcontroller finalizes the hyperparameter set of the network. 

As alluded to earlier, IBM 65nm is used to implement the full-chip, and the power consumption is estimated using Cadence Voltus while clocking the system at 50MHz. Several optimizations are considered when implementing the chip to keep power consumption as low as possible. For example, the random weights of the reservoir are generated using LFSRs rather than being stored in SRAM or DRAM, resulting in a 20.1\% reduction in power consumption. Also, we fulfilled the ESP without the need to compute the eigenvalue, which is supposed to be used later to normalize the reservoir synaptic weight values. Reducing data movement via transferring only the actionable information, i.e. ignoring small activations, played an important role in reducing power consumption. 

When comparing our work with the previous implementations in the literature, listed in \tb{HardwareAnalysis}, one may observe that there is a clear absence of ASIC implementations of ESN network since the majority of previous designs ported to FPGAs. FPGAs provide the necessary reconfigurability, but it always comes at the expense of limited operating speed and high power consumption, which is primarily directed toward interconnected resources~\cite{chen2012nano}. Therefore, due to these reasons and the optimization techniques mentioned above, we have achieved orders of magnitude reduction in power consumption. 


\section{Conclusions}
In this work, a neuromorphic system that is based on ESN is proposed. The proposed system is endowed with neuro- and synaptic plasticity locally without cloud support, ensuring system reliability, fast adaptation, minimum power consumption, and better security. Furthermore, it satisfies the critical conditions required to get the ESN working at the edge of chaos, such as echo state property, in a very efficient manner. The proposed design is benchmarked using time-series datasets to recognize human activities and finger movements. The hardware implementation of ESN has been found to experience a slight degradation in classification accuracy of less than 1.8\%, which is mainly attributed to the limited precision and simplified representation of the activation functions. When evaluating the proposed design for throughput, we observed that the local rings data movement topology may significantly enhance network throughput, but at the expense of $\sim$2\% performance deterioration. Thus, we resorted to MH-Tree, which trades off between throughput and performance. Against power consumption, several optimization techniques at the hardware level are adopted to keep power as minimum as possible and to enable porting of the proposed chip on edge devices with stringent resources.


\bibliographystyle{IEEEtran}
\bibliography{IEEEabrv,refer}

\begin{IEEEbiography}
[{\includegraphics[width=1in,height=1.25in,clip,keepaspectratio]{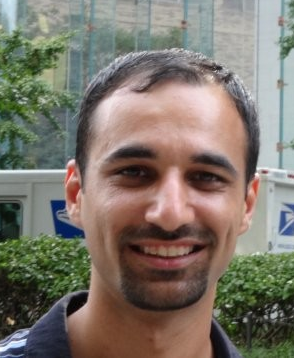}}]{Abdullah M. Zyarah} is a faculty member at the Department of Electrical Engineering, University of Baghdad, and a research scientist at the Department of Electrical and Computer Engineering of the University of Texas at San Antonio. He specializes in digital and mixed-signal designs and his current research interests include neuromorphic architectures for energy constrained platforms and biologically inspired algorithms. Mr. Zyarah received the B.Sc. degree in electrical engineering from the University of Baghdad, Iraq, in 2009 and  the M.Sc. degree in the same discipline from Rochester Institute of Technology, USA, in 2015, where he also received his Ph.D. degree in electrical and computer engineering in 2020. 
\end{IEEEbiography}

\vskip -2\baselineskip plus -1fil
\begin{IEEEbiography}[{\includegraphics[width=1in,height=1.25in,clip,keepaspectratio]{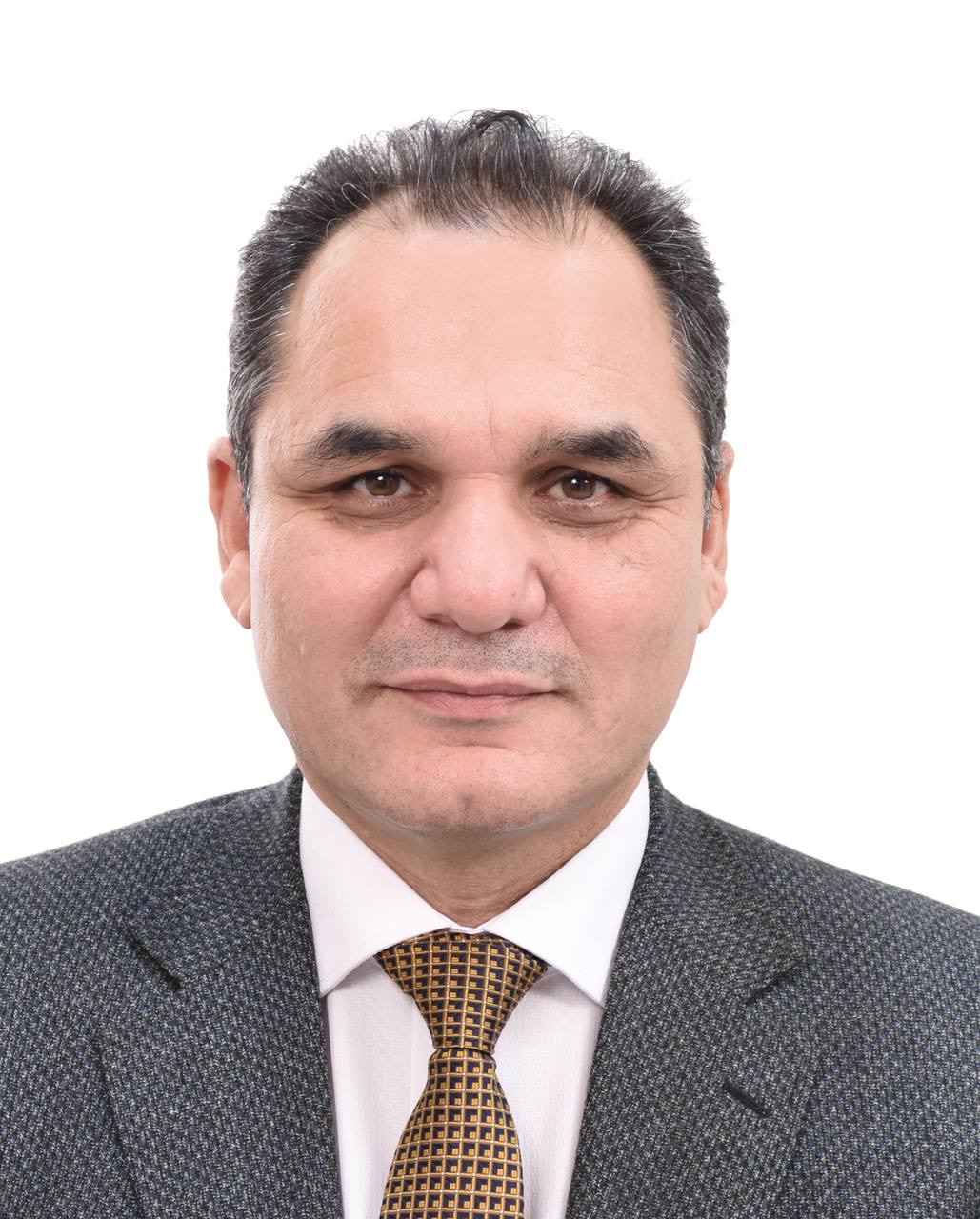}}]{Alaa M. Abdul-Hadi} Alaa M. Abdul-Hadi received his Ph.D. in IT engineering from the National Aerospace  University (Kharkov Aviation Institute), Ukraine, in 2014 and his M.Sc Degree in Computer Engineering from the University of  Baghdad, Iraq, in 2005. He worked as a teaching and research assistant in Horst Gorst Institute for IT Security, Bochum University, Bochum, Germany. Currently, he works as a faculty member in the department of computer engineering, university of Baghdad, Iraq. His research interest includes systems and networks security, trusted computing, AI security, and computer networks.
\end{IEEEbiography}

\vskip -2\baselineskip plus -1fil
\begin{IEEEbiography}
[{\includegraphics[width=1.25in,height=1.2in,clip,keepaspectratio]{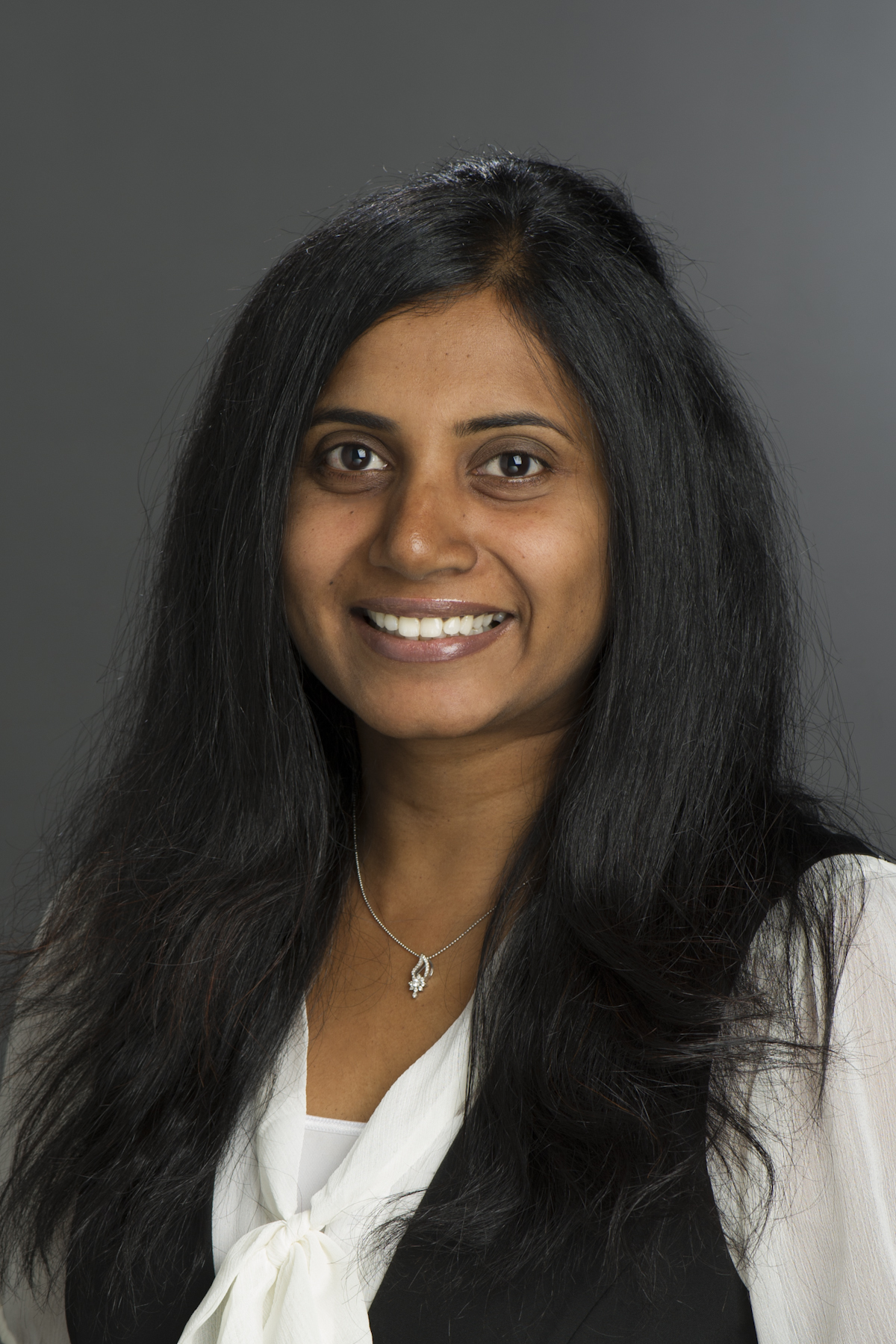}}]{Dr. Dhireesha Kudithipudi}[M'06, SM'16] is a professor and founding Director of the AI Consortium at the University of Texas, San Antonio and Robert F McDermott Chair in Engineering. Her research interests are in neuromorphic AI, low power machine intelligence, brain-inspired accelerators, and use-inspired research. Her team has developed comprehensive neocortex and cerebellum based architectures with nonvolatile memory, hybrid plasticity models, and ultra-low precision architectures. She is passionate about transdisciplinary and inclusive research training in AI fields. She is the recipient of the Clare Booth Luce Scholarship in STEM for women in highered (2018) and the 2018 Technology Women of the Year in Rochester.
\end{IEEEbiography}

\end{document}

%% file: Config.tex
\usepackage{subfigure}
\usepackage{blindtext}
\usepackage{graphicx}
\usepackage{cite}
\usepackage{csquotes}
\usepackage{caption}
\usepackage{subfig}
\usepackage{amssymb}
\usepackage{amsmath}
\usepackage{siunitx}
\usepackage{multicol}

\usepackage{xcolor,colortbl}
\definecolor{Gray}{gray}{0.90}
\newcolumntype{a}{>{\columncolor{Gray}}c}
\usepackage{xcolor,soul}
\definecolor{light-gray}{gray}{0.95}
\sethlcolor{light-gray}

\graphicspath{{Images/}}
\newcommand{\fig}[1]{Fig.~\ref{#1}}
\newcommand{\tb}[1]{Table~\ref{#1}}
\newcommand{\eq}[1]{(\ref{#1})}

\DeclareCaptionLabelSeparator{periodspace}{.\quad}
\usepackage{enumitem,booktabs}
\usepackage[referable]{threeparttablex}
\renewlist{tablenotes}{enumerate}{1}
\makeatletter
\setlist[tablenotes]{label=\tnote{\alph*},ref=\alph*,itemsep=\z@,topsep=\z@skip,partopsep=\z@skip,parsep=\z@,itemindent=\z@,labelsep=.2em,leftmargin=*,align=left,before={\footnotesize}}
\makeatother